# Embedded System Performance Analysis for Implementing a Portable Drowsiness Detection System for Drivers

Minjeong Kim[1], Jimin Koo[2]

[1] Monta Vista High School, Cupertino, CA 95014; minjeongsunnykim@gmail.com
[2] Cupertino High School, Cupertino, CA 95014; jimin.skoo@gmail.com

**Abstract:** Drowsiness on the road is a widespread problem with fatal consequences; thus, a multitude of systems and techniques have been proposed. Among existing methods, Ghoddoosian et al. utilized temporal blinking patterns to detect early signs of drowsiness, but their algorithm was tested only on a powerful desktop computer, which is not practical to apply in a moving vehicle setting. In this paper, we propose an efficient platform to run Ghoddoosian's algorithm, detail the performance tests we ran to determine this platform, and explain our threshold optimization logic. After considering the Jetson Nano and Beelink (Mini PC), we concluded that the Mini PC is most efficient and practical to run our embedded system in a vehicle. To determine this, we ran communication speed tests and evaluated total processing times for inference operations. Based on our experiments, the average total processing time to run the drowsiness detection model was 94.27 ms for the Jetson Nano and 22.73 ms for the Beelink (Mini PC). Considering the portability and power efficiency of each device, along with the processing time results, the Beelink (Mini PC) was determined to be most suitable. Also, we propose a threshold optimization algorithm, which determines whether the driver is drowsy, or alert based on the trade-off between the sensitivity and specificity of the drowsiness detection model. Our study will serve as a crucial next step for drowsiness detection research and its application in vehicles. Through our experiments, we have determined a favorable platform that can run drowsiness detection algorithms in real-time and can be used as a foundation to further advance drowsiness detection research. In doing so, we have bridged the gap between an existing embedded system and its actual implementation in vehicles to bring drowsiness technology a step closer to prevalent real-life implementation.

## 1. Introduction

*1.1 Background*

Drowsiness on the road is a widespread problem with fatal consequences. Each year, 1.35 million people are killed on roadways around the world, according to the World Health Organization's Global Status Report on Road Safety in 2018 [1]. 328,000 drowsy driving-related crashes occur annually in the United States alone, according to a study by the AAA Foundation for Traffic Safety [2]. Furthermore, NHTSA estimates suggest that fatigue-related crashes cost society $109 billion each year, on top of property damage [2]. Studies from the National Sleep Foundation revealed that 50% of U.S. adult drivers admit to consistently driving while drowsy, and 40% admit to falling asleep behind the wheel at least once in their driving careers [3]. Thus, drowsy driving is a ubiquitous problem that claims heavy social and economic tolls. It must be promptly addressed to save irreplaceable human lives and economic damage.

*1.2 Prior work*

Because drowsy driving is a fatal problem with significant consequences, various approaches have been proposed in the field. Some solutions to detect drowsy driving involve the use of biological signals such as electroencephalogram (EEG) data [4–10], optical brain wave signals through functional Near InfraRed Spectroscopy (fNIRS) [11], heart rate variability [4,12], etc. [13]; For example, Arefnezhad et al.'s [8] method measures PERcentage of eyelid CLOSure (PERCLOS), a widely used drowsiness detection metric, using EEG electrode data and a Bayesian filtering solution. In their work, Theta and Delta power bands on the EEG spectrum were identified to be correlated with PERCLOS values. Such methods aim to detect physiological changes in the drivers' biosignals that indicate drowsiness. However, these approaches are not practical for application because the driver has to wear bulky, intricate instruments on their head for extended periods of time, which may be too complex for lay people to use and can impede driving.

Some existing technologies have also detected drowsy driving using driving performance data such as vehicle turning curvature, accelerations, line weaving behaviors, etc. [14,15] For example, built-in systems like the Mercedes Attention Assist system assesses personal driving styles to detect irregularities considering 70 parameters and external factors [16]. Similarly, Castignani et al. [17] proposed the use of existing smartphone sensors to develop a driver profile and detect "risky driving events", taking aspects like route topology and weather into consideration. However, such external factors are more indirect methods for drowsiness detection, as the correlation between driving patterns and drowsiness has not been adequately established. Other studies on drowsiness detection span a wide range of systems and metrics, such as head angle, explicit drowsy signals (yawning, nodding off), blink count and patterns [18–21]. For example, Xu et al. [22] and others proposed systems based on PERCLOS [23,24] and approximated blink statistics, which demonstrated a drowsy behavior-detecting accuracy of over 90%. However, the work focused on detecting simple eye closure rather than predicting drowsiness levels.

Most eye-tracking based drowsiness detection algorithms were based on finding the PERCLOS, as mentioned above, which indicates the percentage of time during which the eye is closed over a duration of time. For example, Feng You et al. [25] proposed a program that detects drowsiness by checking whether eyes are closed using eye aspect ratios calculated from facial landmarks. The program uses a deep-cascaded convolutional neural network to detect the face, then a Dlib facial landmark detector to analyze these facial landmarks. A support vector machine (SVM) was developed to classify if the eyes are open or closed in each frame. If the PERCLOS is greater than a designated threshold, the driver is labeled as drowsy. Most of the research focused on the development of the eye close-open detection model rather than the relationship between eye aspect ratio and drowsiness level. Feng You et al. emphasizes that unlike PERCLOS-80, which predicts that eyes are closed if more than 80% of the pupil is covered by the eyelid, their proposed method measures eye aspect ratios. Using this metric allows the program to account for differences in each user's facial features. Similarly, Manishi et al. [26] proposed a method that uses a Viola-Jones algorithm to detect the eyes and mouth, from which a convolutional neural network is used to detect if the eyes and mouth are open or closed. Closed eyes or wide-open mouths indicate drowsiness. Using the NTHU Drowsy Driver Detection Dataset, this model takes into consideration the different situations that the user may use the program, such as in the dark and with or without glasses.

However, we believe that when there is a detectable difference in the percent of eyelid closure, the driver is already in a dangerous situation. Ideally, drowsiness detection algorithms should "predict" drowsiness in earlier stages by analyzing more subtle signs. Thus, we continued to search for a drowsiness detection algorithm that could detect drowsiness levels while the driver still has enough time to take precautionary steps.

We found that Ghoddoosian et al. [27] proposed a drowsiness detection method that could detect the early signs of drowsiness. Their study also includes a dataset with which we trained the model and tested it ourselves. The proposed LSTM model uses a distinct sequence of blinking patterns such as the speed of eyes closing and opening and the duration, amplitude, and frequency of blinks to predict the early signs of drowsiness. The method detects faces using Dlib's face detection model [28], then uses Kazemi and Sullivan's [29] model to detect facial landmarks. Soukupova and Cech [30]'s model was used to detect blinks.

The LSTM model uses data on a sequence of blinks to classify a person's state as alert, low-vigilant or drowsy. The LSTM model was trained on a dataset of 60 different individuals of a wide range of ages and ethnicities who provided videos of themselves in different states of drowsiness. This algorithm appeared the most promising because it could detect early signs of drowsiness, which is why we decided to use it as a base model for our research. However, this method was developed on a powerful desktop computer, and a platform implemented in embedded systems is needed to run the algorithm in vehicles. Embedded systems use CPUs or GPUs that are less performant than desktop CPUs.

At the same time, while there are existing solutions that implement drowsiness detection algorithms in embedded systems, we have found that the programs are too simple and do not consider the complexity of drowsiness expression or require intrusive methods of data collection. For example Lunbo Xu et al. [22] proposed a smartphone-assisted drowsiness detection system. In their study, Xu et al. validated that all processes of the algorithm run in less than 100 ms/frame, which supports the smartphone's framerate of 10 fps. Xu's research focuses mainly on the development of the eye close-open detection model, and the reported accuracy of the drowsiness detection program measures only the program's ability to determine whether an eye is closed or open. If the PERCLOS exceeds 0.25, the labels the user as drowsy. Similarly, Eddie E. Galarza et al. [31] proposed a drowsiness detection program that runs on an Android-based smartphone. The drowsiness detection program, however, detects various possible behaviors of drowsiness: head swaying, blinking, and looking to the left or right. The reported accuracy of the proposed algorithm considers only the average of the accuracies of detecting each of these behaviors, rather than the state of being drowsiness. Further research exists on drowsiness detecting embedded systems that connect to a separate device. For example, in his review paper on drowsiness detection systems, Anis. Rafid et al. [32], presented several embedded systems including smartphones. Most of them were connected to external devices that had to be worn, such as an EEG System [10] or other devices that collected physiological data. These devices were tested in lab settings, and as mentioned above, the practicality of using these devices in a moving vehicle setting is questionable. Some also used camera's inertia sensors such as accelerometer or gyroscope to detect changes in driver's driving techniques [21,33]. However, as mentioned previously, the correlation between such driving patterns and drowsiness has not been adequately established.

**Error! Reference source not found.** summarizes the methods of prior research. Prior solutions for drowsiness detection on can be categorized by the input signal, such as bio signals, driving patterns, and facial landmarks. The solutions involving facial-landmark detection can be further categorized by studies that detects simple eye closure or PERCLOS, and the one that analyzes various parameters of blinks. We believe that predictiveness of drowsiness detection, or its ability to detect the initial stages of drowsiness through subtle expressions, is the highest in the blink-analyzing method, which has not yet been implemented in an embedded system formfactor.

The ideal drowsiness detection system would be a portable embedded system that can predict drowsiness using the method suggested by [27]. However, we were not able to find such a system yet. To implement such a system, research is needed to understand the capabilities of embedded systems and the design requirements. For example, the computational power to run the algorithm, electrical power consumption, camera resolution, and communication speed are a few of the requirements. We have found existing research that defines the quantitative metrics necessary to run a cyber security application involving a neural network in an embedded system [34,35]. However, up to the author's knowledge, we could not find such research for a drowsiness system which can run powerful drowsiness prediction neural networks.

**Table 1.** Comparison of prior work in terms of predictiveness, direct sign of drowsiness from eyes, and embedded system formfactor. The prior works were categorized by input signals: biological signals, driving patterns, facial landmarks.

| Category of input signals to detect drowsiness | | References | Predictiveness | Direct sign of drowsiness from eyes | Embedded system formfactor |
|---|---|---|---|---|---|
| Biological Signals | | [4–13] | Not clear | No | No |
| Driving Patterns | | [14–16,33,36] | Not clear | No | Yes |
| Facial Landmarks | Simple closed eye detection based | [18] | Too late | Yes | Yes |
| | PERCLOS based | [22–24] | Too late | Yes | Yes |
| | Blink pattern based | [27–30] | Yes | Yes | No |

*1.3 Our approach*

We propose a portable drowsiness detection system consists of a phone and a mini server which can be paired and placed in a vehicle. The phone serves as the camera and touch-display, and the mini server which is plugged into the vehicle's auxiliary power outlet runs the drowsiness detection model in real-time.

Our system uses WiFi to connect the phone and the embedded server locally without connecting to the internet. If our system ran by connecting to the internet, it would have a greater delay (time lag) regardless of the three embedded systems, because there would be more routers involved between the client (phone) and the server. However, even if we used the internet, the overall round-trip delay would be in the hundreds of milliseconds, which would not be problematic for predicting drowsiness. This is because the throughput performance, rather than the delay, needs to be more enforced in following the 33ms limit. We carefully decided not to use the internet because of privacy concerns, as we believe that users would not want their videos or drowsiness statistics to be accessible through any external server. Furthermore, the cost of LTE communication to the internet from the vehicle can be costly, which makes it less favorable for users.

Because an average vehicle's power outlet (also known as cigarette outlet) only puts out about 120 watts of power, we had to find a device for the server that runs with around or less than 120 watts of power.

We also propose a threshold finder and drowsiness value voting algorithm. We put our research efforts into the following two parts: developing a portable embedded system for performing drowsiness detection and developing an algorithm that finds the threshold for each model and votes for the final drowsiness value. For the drowsiness detection model, we used the same drowsiness detection model as proposed by Ghoddoosian et al. [27]. Our proposed approach not only considers the temporal aspect of drowsiness detection but also allows the model to be utilized in a real-time vehicle setting.

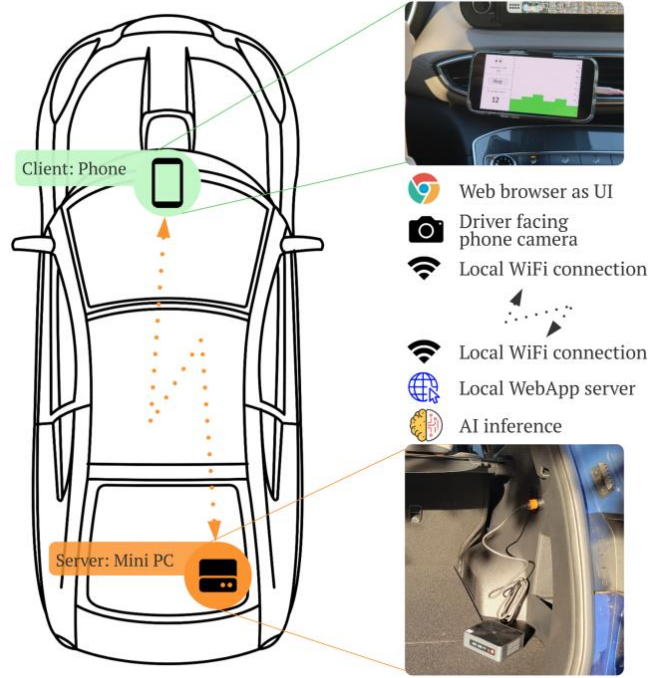

**Figure 1.** Vehicle installation example of our proposed solution. The client is a phone that can be attached near the dashboard area. The front facing camera should see drivers face to detect blinks. The server can be positioned anywhere in the vehicle as long as 12V power can be provided. The client and server is communicating wirelessly through Wi-Fi connection

1.3.1 Portable Embedded System

In this paper, we also propose the hardware/software implementation of a portable drowsiness detection system that can be used in the vehicle. In the user's vehicle, a powerful Mini PC server is set up, and the phone connects to the server and run the drowsiness detection web app. This approach also ensures privacy since the user's data stays in the mini server instead of being shared on the internet. In order to develop our system, quantitative analysis of the communication speed and of the performance of the system in a portable setup needs to be performed. The analysis allows us to decide which of the candidate hardware should be used for our proposed system. We tested the candidate systems and selected the device that can both accommodate the complex AI performed for drowsiness detection and be conveniently placed in a vehicle. As shown in Figure 1, the client can be placed near the front of the vehicle where the camera of the client can stably detect the driver's face. The server can be placed anywhere in the vehicle. For example, it can be placed in the trunk and use the 2V power outlet as its source of power.

1.3.2 Threshold Optimization Algorithm and Voting Algorithm

Furthermore, we propose a method for optimizing the threshold for our model, which divides blink sequences into the following categories: alert or drowsy. The drowsy category includes both the "low vigilant" and "drowsy" categories from Ghoddoosian et al.'s model. We hypothesize that on the road, our drowsiness detection model requires a trade-off between sensitivity and specificity. Sensitivity refers to the ability of the model to accurately detect drowsiness when the driver is drowsy, while specificity refers to the ability of the model to correctly detect alertness when the driver is not drowsy. We decided that the sensitivity of our model is much more important than specificity, as poor specificity may cause inconvenience by falsely alerting the driver, but poor sensitivity can lead to life-threatening situations. Currently, most AI-related papers aim to optimize parameters to maximize the F1 score of their models, which is the percentage of the sum of true positive and true negative predictions. However, by changing some parameters, we can control the trade-off between the false positives (FPs) and false negatives (FNs) and emphasize sensitivity over specificity. Additionally, we propose a voting algorithm, which finds the weighted average of various models' predictions based on the true negative and true positive rates of each of the models.

*1.4 Review of the research process and outline of this paper*

Figure 2 describes a detailed process for the research we have conducted. First, by analyzing prior research and investigating web app technology, we designed an embedded-system architecture that can predict drowsiness levels in real-time in vehicles. We implemented our drowsiness detection algorithm in a Desktop PC before testing candidate embedded systems to check the functionality of each.

We explored possible embedded system candidates that can run our system and developed methods to check the performance of candidate embedded systems. Specifically, we checked the communication throughput to ensure that our system can run without accumulating lag, and also checked processing speeds to process the video streams for detecting drowsiness without lagging. Critical processes were identified: face detection, landmark detection, and blink detection. These will be explained in detail in the sections that follow. From these results, we were able to choose the most practical device for our embedded system. Then, we developed a threshold optimization method that classifies the drowsiness level into certain drowsiness states. The threshold optimization is performed based on the false positive/false negative rates of the models. All of our findings provide the necessary information to finalize the proposed drowsiness detection system for drivers in vehicles.

The remainder of the paper is organized as follows. Section 2 describes our hardware and software setups that explain the algorithms we used to develop and analyze our candidate embedded systems' performance. Section 3 describes how we ran these procedures to validate our system's performance and our Threshold Optimization Algorithm. Section 4 explains and analyzes the specific results from the collected data, and lastly, Section 5 concludes our study.

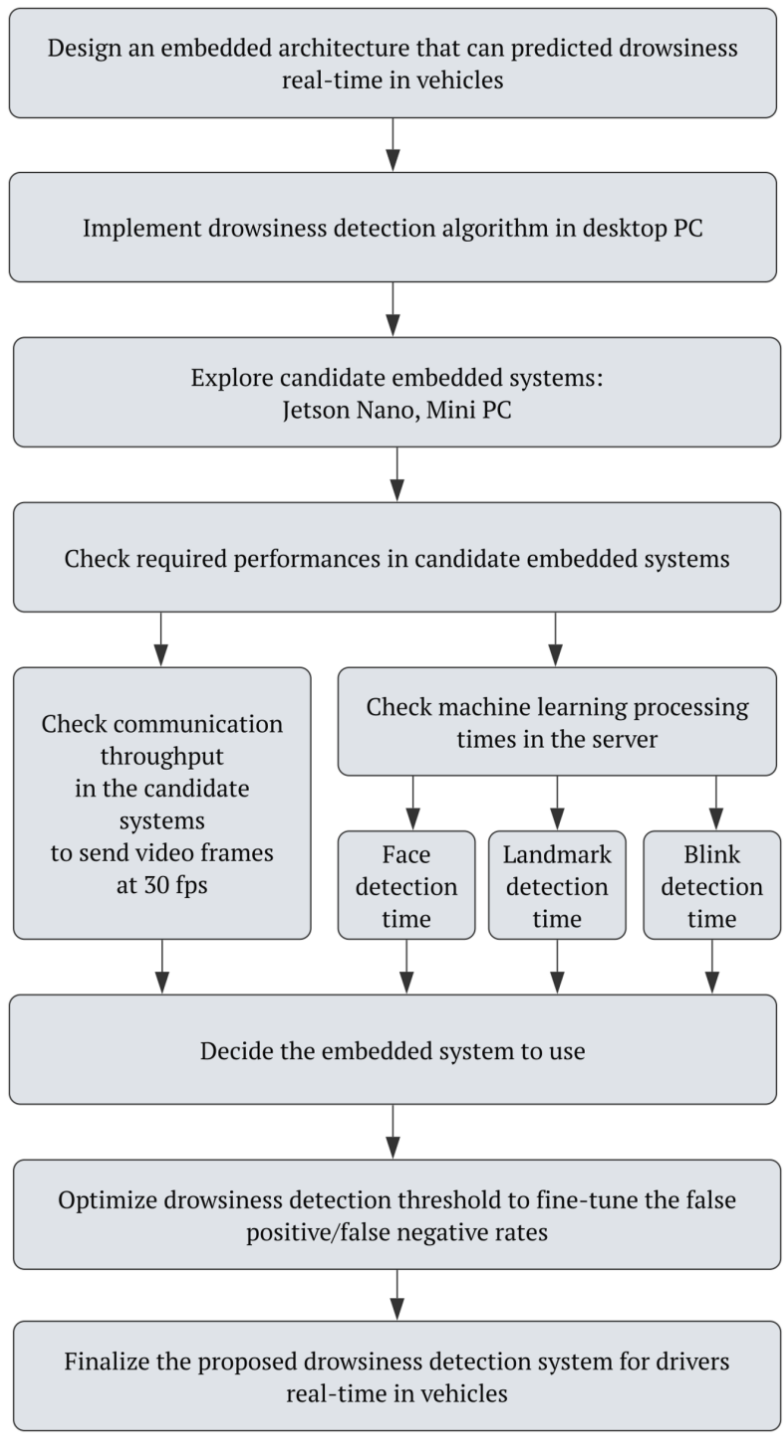

**Figure 2**. Overall research process to develop proposed drowsiness detection embedded system in vehicles.

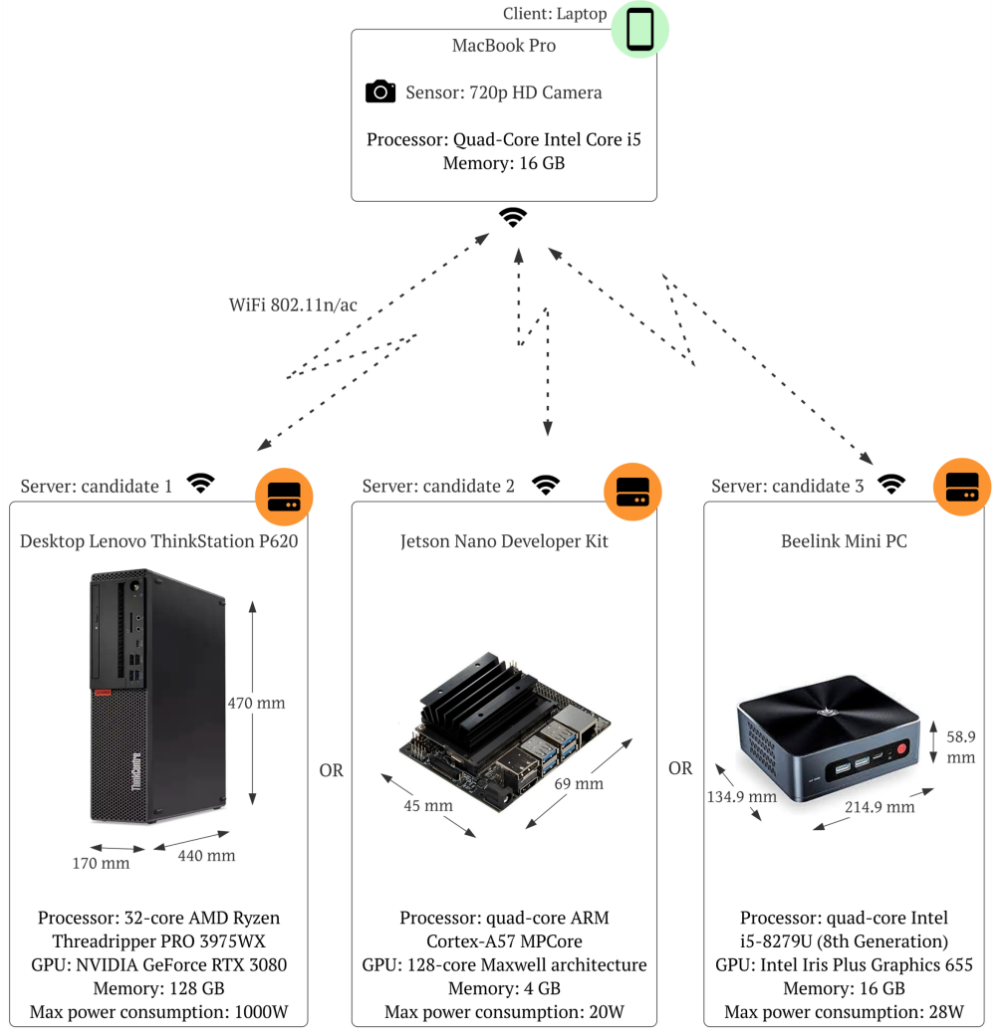

**Figure 3.** HW Setups for experiments. MacBook Pro was used as a client instead of a phone and three devices are listed as servers: Desktop PC, Jetson Nano development kit, Mini PC. One of the candidate servers was powered up to run the drowsiness detection system. The dimension and specification of the candidate servers are described below each device.

**2. Materials**

In this section, we describe the hardware and software configurations of our proposed embedded system that runs the drowsiness detection algorithm in real time, including the drowsiness detection algorithms, threshold optimization method, and the real-time voting algorithm to calculate the final drowsiness value.

*2.1 Hardware Setup*

As described earlier, we are proposing a drowsiness detection system that uses two parts to detect drowsiness: a client and a server. The client can be any mobile phone that has a camera and can connect to Wi-Fi. The phone should also have a web browser that can load our web app. The server needs to be powerful enough to run the web server and drowsiness detection algorithm, which means that it needs to have adequate communication performance and processing power.

To measure the performance of the communication and server's processing time, we used a laptop as our client because it provides better tools and an ideal SW environment for collecting data from the web app client. The laptop's default camera (supporting 720p image resolution and 30 fps maximum framerate) was used as the client camera to take video and stream to the server. Since most phones also support fast WiFi connection and have high resolution cameras, the client side can easily be replaced by a mobile phone.

We chose to investigate a Jetson Nano Developer Kit and a Beelink Mini PC as potential embedded systems based on their performance, power efficiency, and formfactor. The specifications for each of the devices used are detailed in Figure 3.

We selected the Jetson Nano as one of our candidates because of its GPU-accelerated performance and efficient power consumption. The Nano contains a quad-core Cortex-A57 processor, which makes it similar to other embedded systems. However, it also has 128 Maxwell architecture-based GPU cores, which accelerate the AI interference process. While there are many other embedded systems such as Raspberry Pis, Arduinos, or BeagleBoards, these devices do not have powerful GPUs or neural network accelerators to perform machine learning processes efficiently. Suzen et al. [37]compared the performances of NVidia's TX, Jetson Nano, and Raspberry Pi 4 Model B (the newest Raspberry Pi model) by running a Deep-CNN that classifies clothing images into 13 categories. They found that the Jetson Nano could run inference operations more than five times faster than a Raspberry Pi. The Jetson Nano also only uses 10 W of power, which is less than the 120 W output of a typical vehicle's power outlet.

Our other candidate was the Beelink Mini PC, which we selected because of its high-performance CPU and GPU, which are similar to those of a typical laptop performance. The Mini PC has an 8th generation Intel i5-8279U process and typical Intel GPU. The power consumption can be up to 28 W, which is still less than the 120 W output of a typical vehicle's power outlet.

Before implementing the embedded server system, we used a desktop PC with AMD Ryzen Threadripper PRO processor. Although the PC is not in an embedded system formfactor, we used it to develop our system to train the model and compile the web-server code. The PC has a 32-core AMD Ryzen Threadripper processor with a powerful NVDIA GeForce RTX 3080 GPU. After developing the code in this machine, we transferred the package into a Docker image and ran it on the Jetson Nano and Mini PC.

When running the experiment, we connected the client and server via WiFi 802.11 n/ac connection. Both devices were in the same room and only one of the servers was turned on to run the web server. The server had a separate local monitor and keyboard/mouse to control the web server status.

When running in vehicles, the phone can be held by an aftermarket phone holder and connected to a USB port in a car. The server can be connected to the 12V power outlet via a 12V to 19V DC converter, and installed in any secure place that does not shift significantly when the car is in motion.

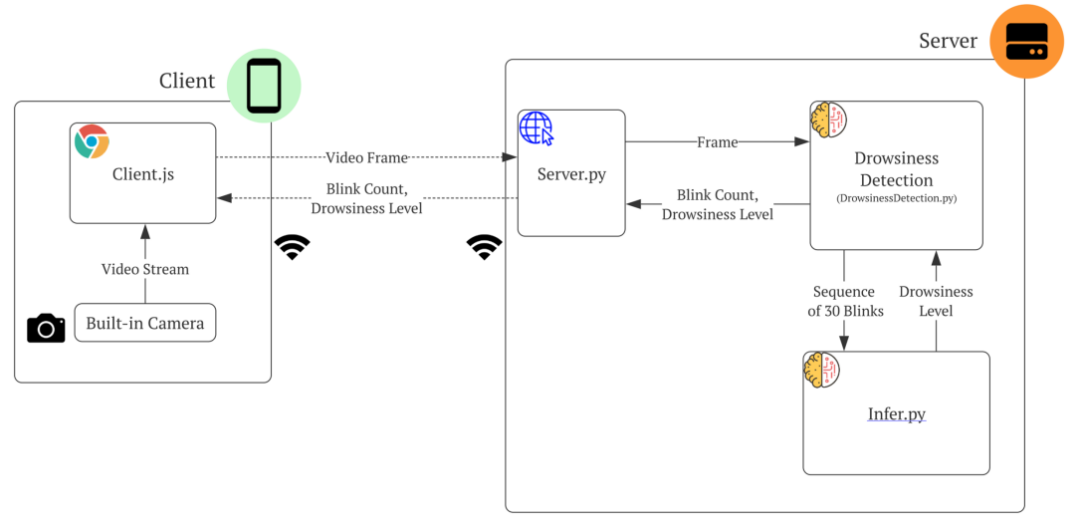

**Figure 4.** Proposed drowsiness detection SW architecture that runs the web app interface. Major block diagrams are shown for the client and server side.

*2.2 Software*

The overall architecture and block diagrams of our web app's software are shown in **Error! Reference source not found.**. The software consists mainly of two parts: the client side and the server side, which we will explain below.

2.2.1 Client

Our system was built to be compatible with phones: the phone functions form the "client" part, utilizing the camera and screen to communicate with users, while the server functions form the computational device. When the client connects to the server, it automatically downloads the files needed to run the web app, which include index.html, client.js, and base.css. These files are provided in the AioRTC server example. Index.html provides the input interface to the user and displays the elements that users interact with as in **Error! Reference source not found.**, while client.js includes the majority of the functionality of the web app, such as video display, WebRTC communication, display of the drowsiness detection level, and history of the drowsiness levels in real-time. Client.js receive a video stream from the built-in camera, from which is sends frames to the server, as shown in the Client side of Figure 4. The CSS file contains the styling elements, which include color schemes and the formatting of the main page. Currently, we are using a simple method of alerting drivers by displaying their drowsiness value and changing the color of a bar graph to red when the user's drowsiness level passes a certain threshold.

2.2.2 Server

For our server, we utilized AioRTC (https://github.com/aiortc/aiortc), which is an open source library for Web Real-Time Communication (WebRTC) and Object Real-Time Communication (ORTC) in Python. We chose to use AioRTC because its implementation is simple and readable compared to WebRTC. In addition, our code handles a local server, meaning there is required communication between the client and the backend (models, server.py). AioRTC was most suitable because it can send data from the 'client' to the 'server' without having to download data locally before analyzing it, ultimately reducing lag. When the device begins detecting drowsiness upon the driver's request, the client (Client.js) sends video frames to the Server.py. We modified server.py to include a section that sends and receives data from the drowsiness prediction server, as shown in the Server side of Figure 4. First, the server sends a fixed 2D array of the video frame to the drowsiness prediction file, DrowsinessDetection.py. DrowsinessDetection.py uses the RNN drowsiness detection model to predict and return drowsiness values along with the total number of collected blinks to server.py. The client receives the information and displays it on the HTML page. We explain DrowsinessDetection.py in more detail in section 2.3.1.

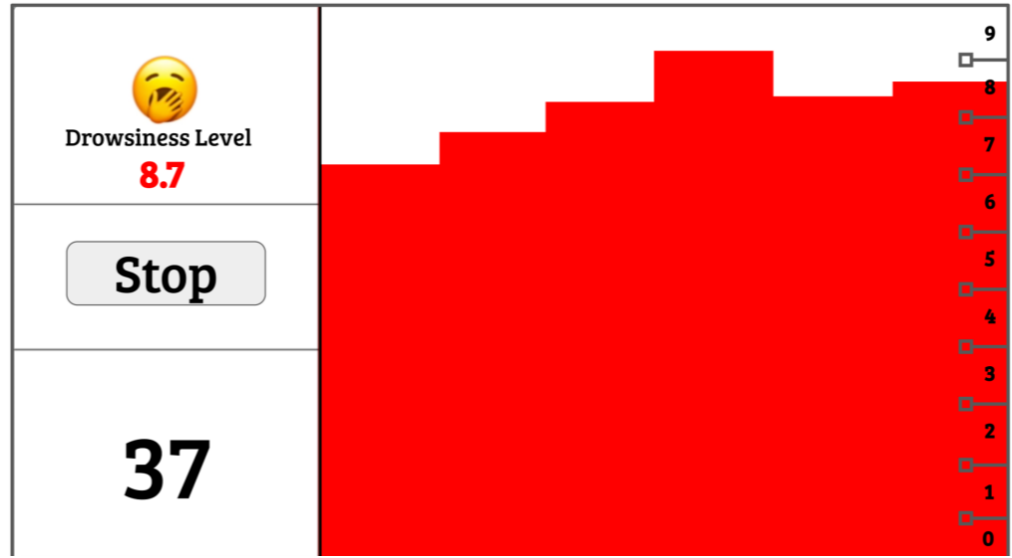

**Figure 5.** Running example of the user interface. The emoji on the top left describes the current drowsiness status and the bar chart displays on the right the historical drowsiness value over time. On the bottom left, the number of detected blinks is shown so that users can know whether the algorithm is running without error.

*2.3 Drowsiness Detection Model*

In the following sections, we summarize the algorithms used and developed by Ghoddoosian et al. for the drowsiness detection model, as well as our implementation that enables our proposed program to monitor the driver in real-time in the server client setup. We utilized Ghoddoosian et al.'s blink detection and feature extraction algorithms and offline training process making minimal changes to the code. For the live monitoring step, we modified the blink feature extraction code to include a function that sends data and receives a drowsiness value. This process is further explained in section 2.3.3. Furthermore, we developed a Threshold optimization algorithm and Voting Algorithm, which both serve to calculate the final real-time drowsiness prediction.

2.3.1 Blink Detection and Blink Feature Extraction

For our blink detection and blink feature extraction steps, we used the models proposed in Ghoddoosian et al.'s paper. We cloned their source code published in GitHub and combined it with the AioRTC webserver example. Their blink detection process can be divided into three steps: face detection, facial landmark detection, and blink detection. First, the program uses Dlib's pre-trained face detector, which is based on the standard Histogram of Oriented Gradients + Linear SVM method for object detection [28]. Ghoddoosian et al. then used Kazemi and Sullivan's [28] facial landmark detector because it was trained with an "in-the-wild dataset" which included videos filmed in various conditions (illuminations, facial expressions, head positions, rotations, etc.), thus making the model robust to a variety of environmental conditions [27]. For the blink detection step, Ghoddoosian et al. used Soukupova and Cech's blink detection model [30] to perform the first blink detection step. Once a blink is detected, the blink retrieval algorithm uses the eye aspect ratios of the eyes to extract four features of the blink: amplitude, velocity, frequency, and duration. For further explanation of these values and their formulas, we refer readers to [27]. Each of these features is then normalized for each individual according to one-third of the blink features extracted from their alert video. This step is significant for this model as all the data are trained together, so differences across each individual's blinking pattern must be accounted for.

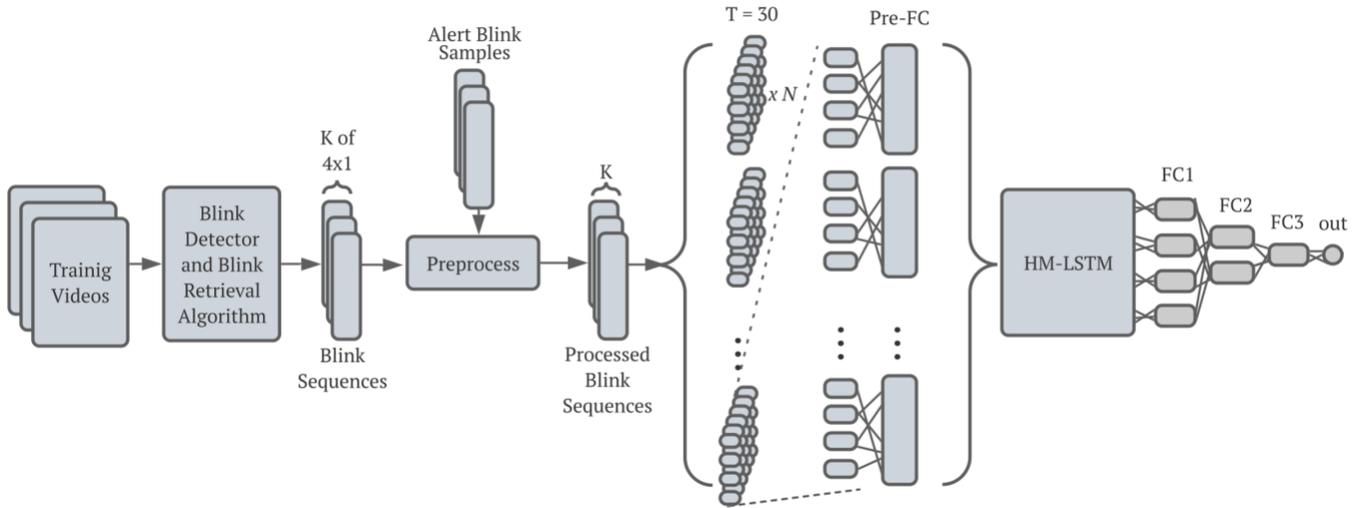

**Figure 6.** Drowsiness detection algorithm suggested by Ghoddoosian et al.

2.3.2 Offline Training and Threshold Optimization Algorithm

Regarding the drowsiness detection model, we used the model proposed by Ghoddoosian et al. The structure of the model is shown in Figure 6 from [27]. Ghoddoosian et al. introduced an HM-LSTM network to incorporate the temporal element of drowsiness detection, since the Hidden Markov Model (HMM) from [27] indicates that expressions of drowsiness follow a temporal pattern. The model was trained on a dataset of 180 10-minute recordings, totaling to 30 hours of RGB videos. 60 participants were asked to film themselves in three different drowsiness states: alert, low-vigilance, and drowsy [27]. Of the 60 participants, 51 were men and 9 were women, some with facial hair and some with glasses, and of five different ethnicities. Participants were given instructions on how to film their videos; specifically, the phone should have been about an arm's length away from the user and in the location/angle representing that of an actual vehicle setting (suggested placing the phone on computer screen).

This dataset was divided into five separate folders, referred to as "folds," each consisting of 36 videos (3 videos each from 12 participants). During the cross-validation step, a new model was created and trained with videos from four of these folders and tested with those of the remaining folder.

After the blinks of each of these videos are analyzed according to the algorithm explained in the previous section, the values are normalized and passed in as input for the HM-LSTM model. For details on the dataset with which the model was trained and the features of this HM-LSTM network, we refer readers to [27].

In Ghoddoosian et al's paper, training.py trains the model by classifying each blink sequence's predicted drowsiness value, a number from 0.0 to 10.0, into three categories: alert, low-vigilance, and drowsy. Then, it compares the predicted category of blinks to the label of these blinks. The blinks are classified according to the ranges listed below.

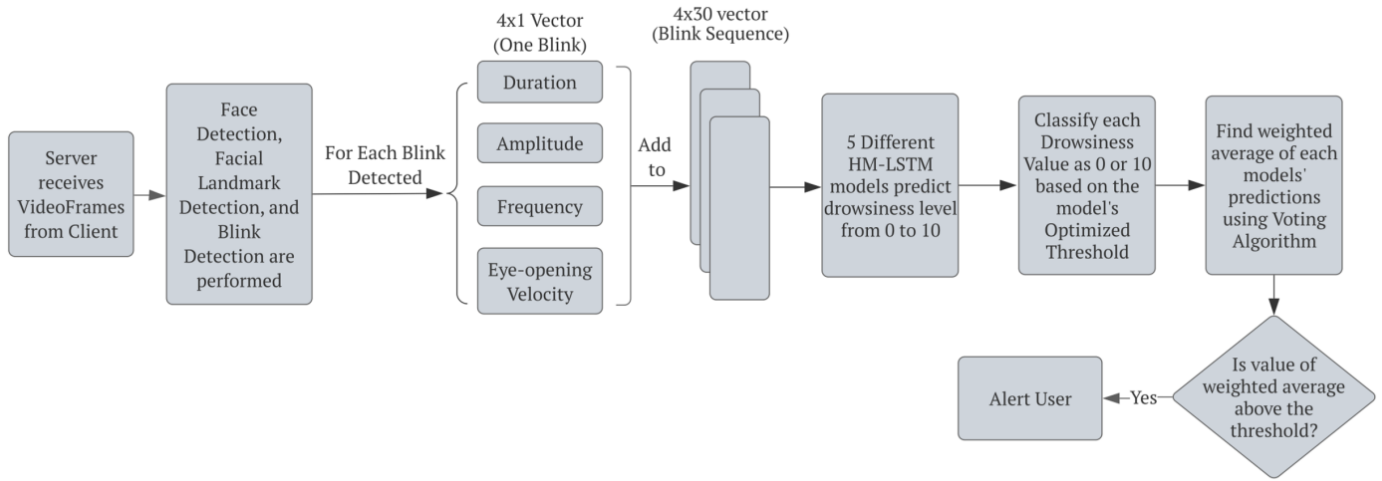

**Figure 7**. Real-time drowsiness detection and alerting algorithm running in server

- Alert: 0.0 predicted value < 3.3
- Low vigilance: 3.3 predicted value 6.6
- Drowsy: 6.6 <predicted value 10.0

For our proposed solution for drowsiness detection for drivers, we added an additional step after the model training. First, we group the low-vigilance and drowsy categories together since we want to alert the driver once they begin expressing any signs of drowsiness. Also, we do not fix the threshold to one value (i.e. 3.33) as Ghoddoosian's training algorithm does. Instead, we developed a customized threshold value which is determined by the ratio of false positive to false negative values of each model's confusion matrix. False negative (FN) rates indicate the fraction of predictions that incorrectly predicted a blink sequence as drowsy, while false positive (FP) rates indicate the fraction of predictions that incorrectly predicted a blink sequence as not drowsy. Both cases negatively impact the user experience, but we believe that decreasing the false negative rate is more important than lowering the false positive rate because safety is more important than convenience. We intentionally alter the threshold to perform the trade-off between the values. These rates change according to the threshold since the threshold determines whether a blink is predicted as drowsy or not drowsy. The process of finding the optimal threshold value will be explained in Methods Section 3.3. To test our optimization algorithm, we used the same dataset as the one used by Ghoddoosian et al.

2.3.3 Online Monitoring and Voting Algorithm

The HM-LSTM model proposed in [27] is trained and runs only off-line after all data has been collected. 10-minute videos from 60 people were used to train and test the model. However, an online monitoring algorithm needs to be developed to predict the user's drowsiness level in real-time.

Figure 7 shows the sequence of our online monitoring algorithm. We implemented it in two main files: DrowsinessDetection.py and Infer.py. We developed DrowsinessDetection.py based on the blinkvideo.py, which Ghoddoosian et al. uses to extract blink features for the model input. Instead of saving the blink features as a text file, however, DrowsinessDetection.py calls a function in Infer.py using a list of these values as the parameter. Infer.py then uses TensorFlow to run a session for each of the drowsiness detection models and get the drowsiness value. DrowsinessDetection.py returns the drowsiness value and number of blinks collected back to the AioRTC server file, as mentioned in section 2.2.2.

To predict the final drowsiness level, we developed a voting algorithm that find the weighted average of the predicted drowsiness values from each of the 5 models. Notably, Ghoddoosian et al. validated the drowsiness detection algorithm using a leave-one-out cross-validation step, which results in five different drowsiness detection models trained with varying combinations of datasets. Specifically, the training dataset was split into five different groups; the first model was trained using groups 2,3,4 and 5 and tested using group 1, the second model was trained using groups 1,3,4,5 and tested using group 2, etc. For further explanation of these models, we refer readers to [27]. Each model outputs a drowsiness prediction value, which is categorized according to the thresholds found by the Threshold Optimization Algorithm. Alert blink sequences are assigned the value 0.0, and drowsy blink sequences are assigned the value 10.0. To obtain the final drowsiness value from the values predicted by the five models, our Voting Algorithm which finds the final value from their weighted average. Each model is assigned a weight value Vi, which is the sum of the true negative rate of the model and double the true positive rate for model i. We assigned a greater coefficient to the true positive rate than the true negative rate because we value sensitivity over specificity as explained in the introduction.

$$V_i = 2TP_i + TN_i \qquad (1)$$

We then take the sum, *S*, of all $V_i$ across models. S will be used to normalize the weighted values, as shown below.

$$S = \sum_{i=1}^{N} V_i \qquad (2)$$

Finally, we define the prediction value, *P*, as the final predicted drowsiness value as shown below.

$$P = \sum_{i=1}^{N} \frac{V_i}{S} b_i \qquad (3)$$

$b_i$ is the binary value indicating whether the driver is drowsy or alert (1 if drowsy, 0 otherwise). If the value *P* surpasses 0.5, we predict the driver is drowsy, and the device alerts the user. The value *P* is displayed on the client's screen in the form of a bar chart, as illustrated in Figure 5.

## 3. Methods

We proposed a portable embedded solution for a drowsiness detection system that can be mounted on a vehicle. In order to validate our solution, we decided to check the following characteristics: communication throughput between the client and the server, processing time to detect the face, facial landmarks, and blinks, and the algorithm for finding optimized threshold value based on the false positive and false negative rates of the drowsiness detection models. In the following sections, we describe the detailed measurement procedures to collect the necessary data and the analysis we performed for each of the items above.

*3.1 Communication Speed*

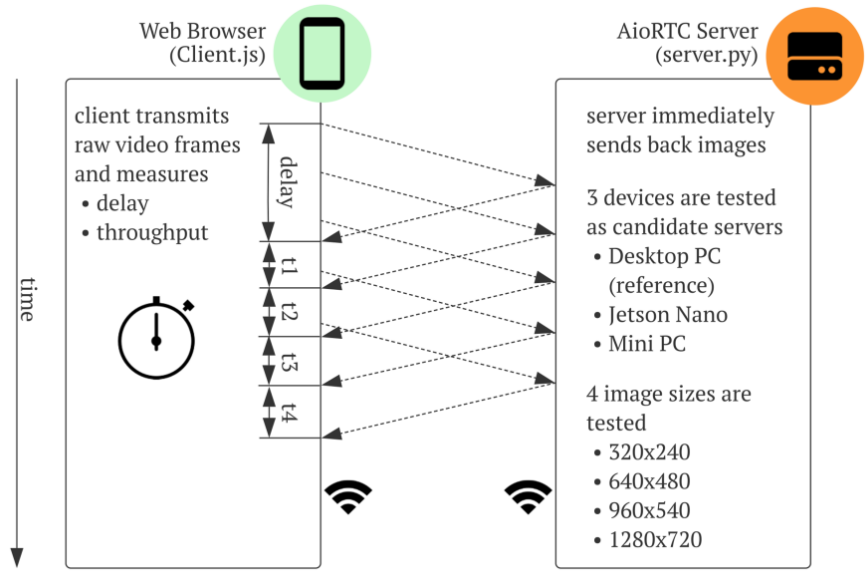

**Figure 8**. Method for measuring communication speed between client and server. The delay time is defined as the time for a video frame to send to server and received back to client. The through is defined as the number frames that can be transferred per one second. We inserted a code to check the arrival time of the frames and measured the time difference between frames to measure the throughput.

Our proposed system requires fast enough communication speed for the drowsiness system to work properly. If the speed is not fast enough to transmit video frame rate from client to server, the client will start to accumulate unsent video frames causing delays and eventually a memory overflow and system crash.

Communication speed can be explained in terms of delay and throughput. The delay between the client and server determines the amount of time it takes for a frame sent from the client to be accessed by the server. The throughput is defined as the amount of data that can be sent in a given time unit. In our case, we used frames per second (fps) as the unit. In **Figure 8**, the delay is described as a video frame that is sent from the client to the server and sent back to the client. The throughput is expressed in terms of time intervals (t1, t2, t3, t4), which is the inverse of the throughput. In our implementation, frames must be sent to the server at a rate equal to the camera's frame rate in order for the system to function in real-time without accumulating lag. On the other hand, the communication delay, which is the round-trip travel time of sending a frame from the server and receiving it, has a negligible effect on the efficiency of the proposed system since it does not cause lag to accumulate. An increase in the delay time would simply cause a delay in getting the drowsiness value displayed on the client's screen. Since delay time is typically in the order of a few tens of milliseconds for WiFi, its effect will be negligible. Thus, we only analyzed the throughput of our system.

To evaluate the communication throughput of our proposed system, we set up the AioRTC server to send back the original video frames and modified the client-side code (client.js) to record the time when the client receives the returned frame from the server, which is described as t1,t2, t3, t4. We collected 500 data points in the client web browser and downloaded the collected data for processing later. Both the delay and throughput were measured, but only the throughput was analyzed in the results section.

Furthermore, in order to see whether the communication throughput supports different

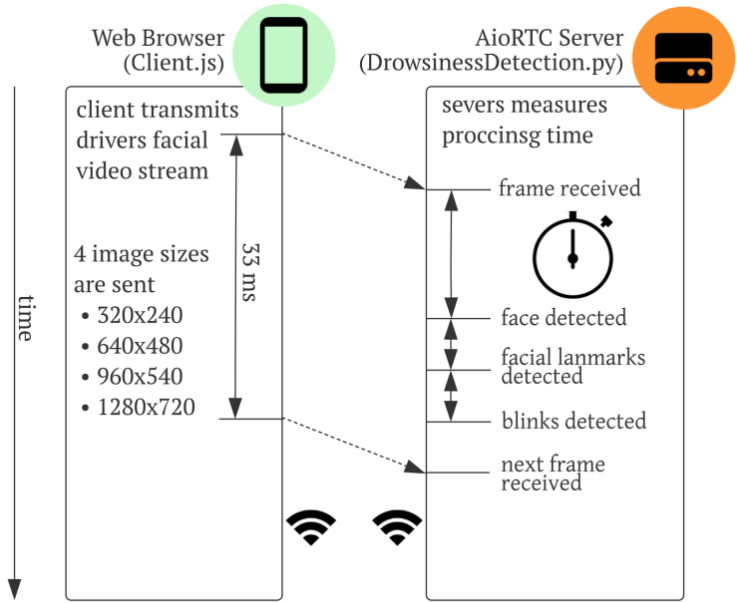

**Figure 9**. The timestamps which were used to analyze the processing time of drowsiness detection algorithm: the time required to detect face, detect facial landmarks, detect blinks are logged

image resolutions, we sent four different image resolutions 320x240, 640x680, 960x540, and 1280x720. If the communication throughput is not high enough, certain video resolutions may not be supported and cause accumulating lag over time.

We measured the communication throughput of the three candidate devices for servers separately.

*3.2 Processing Time*

In order to implement a drowsiness detection system in an embedded system inside a vehicle, we needed to make sure the server is capable of processing the algorithm in real-time. For a video with a frame rate of 30 fps, the processing needs to be completed within a 33 ms period. To calculate the processing time, we added a few lines of code that print out timestamps in our full software, which includes the AioRTC server and drowsiness detection algorithm. Only some steps of the drowsiness detection algorithms are run for every frame; those steps are time-sensitive, because if a total processing time per frame exceeds 33ms, the unprocessed frames will start to accumulate on the server side. The processing steps that need to be run for every frame are face detection (Dlib), landmark detection (68-point landmark, OpenCV), and blink detection (SVM classifier explained in [28]). The setup to measure the processing time is shown in Figure 9.

There are also steps in the drowsiness detection algorithm that are less time-sensitive. For example, the drowsiness prediction step using the HM-LSTM model [27] only needs to be performed after each blink. As a person can typically blink up to 20 times a minute, up to a few seconds is allowed for this step giving room for a relatively long delay time for inference. Thus, we did not measure the inference time for the HM-LSTM model.

We measured the processing time by printing timestamps at the following events in the server code, DrowsinesDetection.py:
- Right after the frame is received from the client
- Right after the face is detected (face detection)
- Right after 68 facial landmarks are detected (landmark detection)

- Right after blink detection is performed, which checks if a blink is occurring in the frame (blink detection)

By finding the differences between these logged timestamps, we calculated the net processing time for each step. The same process was done for each of our candidate platforms at four different frame resolutions:
- Devices: Desktop PC, Jetson Nano, Mini PC
- Resolutions:   320x240, 640x680, 960x540, and 1280x720

For each measurement, 450 data points were collected, and the mean and standard deviation values were calculated.

*3.3 Threshold Optimization Algorithm*

In order to find the optimal threshold, we first need to see how different thresholds affects the FN, FP, and 2*FN+FP values. Furthermore, we wanted to compare how those values change when we use the optimal threshold versus the default threshold. Ideally, the optimized threshold would produce a decreased FN rate at the expense of an increased FP rate, which would indicate that the algorithm will correctly detect drowsiness more frequently while giving false alarms more frequently as well. Since Ghoddoosian et al. [27] divided their dataset into five folds, resulting in five different drowsiness detection models created in the cross-validation step, we calculated the FN, FP, and 2*FN+FP values and tested the Threshold Optimization Algorithm for each of the models. The following process was performed for each model.

We propose a method to sweep through the various thresholds and find the threshold that produces the minimum 2FN+FP value, as shown in Figure 10. Since Ghoddoosian et al., used a threshold value of 3.33 to divide the drowsy and low vigilant blinks, we sweep our threshold for dividing drowsy and not drowsy blinks from 3.33 to 10 in increments of 0.33 (21 total thresholds). To evaluate the FP and FN rate of each threshold, we created a function that calculates the values of the confusion matrix for the given threshold value, model, and dataset as input. From the confusion matrix, we get the FP and FN metrics, and calculate the value of 2FN+ FP. We plotted each of these values in a line graph to observe the general trend of these values according to the varying threshold. We used the Pandas method idxmin() to find the threshold producing the minimum 2FN+FP value. After finding the optimal threshold value

```
Algorithm 1 Threshold Optimization Algorithm
 1: Input:
 2: O: array of loaded models
 3: B: array of loaded blink sequences
 4: L: array of loaded blink sequences' labels (0: not drowsy, 10: drowsy)
 5: Output: T: array of optimized threshold value for each model
 6: M ← length(O)
 7: N ← length(B)
 8: for m = 1, 2, ..., M do
 9:     initialize map V
10:     for t = 3.3̇, 3.6̇, ..., 10 do              ▷ iterate through thresholds
11:         initialize array C of length N
12:         for i = 1, 2, ..., N do                ▷ iterate through blink sequences
13:             P = output of model O[m] on blink sequence B[i]
14:             if P ≥ threshold then             ▷ P is a value from 0 to 10
15:                 C[i] ← 10                      ▷ classify blink sequence as drowsy
16:             else
17:                 C[i] ← 0                       ▷ classify blink sequence as not drowsy
18:             end if
19:         end for
20:         Calculate confusion matrix by comparing C and L
21:         Calculate FN and FP from confusion matrix
22:         V(t) ← 2FN + FP
23:     end for
24:     T[m] = argmin V(t)                          ▷ find optimal threshold for model O[m]
             t
25: end for
```

**Figure 10**. Pseudo code for the Threshold Optimization Algorithm. The algorithm finds the optimal threshold value which minimizes the value of 2FN+FP.

and the pertaining FN, FP, and 2*FN+FP values, we compared them with the same values produced by the default threshold (6.67).

**4. Results and Discussion**

We successfully implemented the proposed client/server platform to predict drowsiness for drivers in vehicles. The implemented code functioned desirably, providing the drowsiness level and classifying the drowsiness state. The platform serves as a web app from a local server so that any client device, such as a phone or laptop, can launch the web app. However, since the server needs to be powerful enough to run AI models that detect drowsiness levels from the raw camera images, we developed the methods to validate the performance of candidate server devices by measuring the processing time. Additionally, since the communication speed needs to be fast enough to transmit the video frame data and processing speed, we developed the methods to measure each of the quantities above. In the following section, we present the results for each candidate device in this section.

The platform we proposed can be used by other researchers who want to develop their own drowsiness prediction algorithms, as they can install and test their systems in their vehicles and check the communication speed and processing time as we have shown. In Section 4.1, we describe how we checked the communication speed, and in Section 4.2, we describe how we measured the processing time of our candidate servers. Detailed interpretations of the results are shown.

We successfully calculated the effect of changing the thresholds on the FN and FP rates of our models and optimized the threshold based on the value of 2FN+FP. We compared the effectiveness of the optimized threshold values by comparing the FP, FN values calculated from the optimized threshold with those of the default threshold. This approach can also be used for other applications where a trade-off between specificity and sensitivity needs to be made. Section 4.3 explains the details of the calculated results and the analysis.

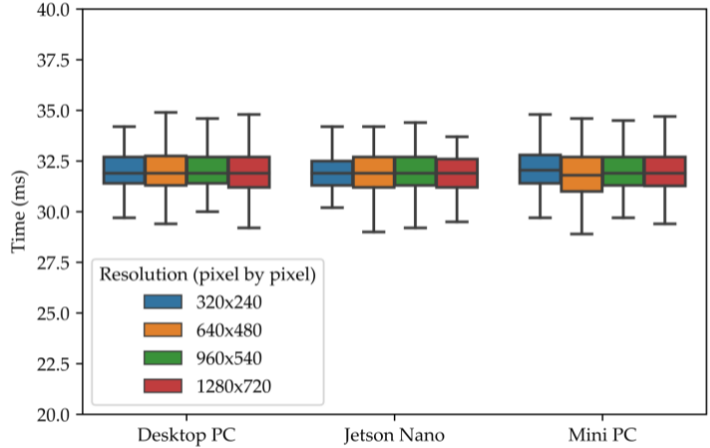

**Figure 11.** The time intervals between frames when a video stream is sent from client to server and then received back. Regardless of video frame resolutions or server platforms, the time intervals were around 33 ms which corresponds to 30 fps. Small variations (±4.154 ms) existed from the mean value between frames.

*4.1 Communication Speed*

As shown in Figure 11, the median time difference between frames being received by the client side was consistently close to 33.3 ms for all three cases: the computers, Jetson Nano, and Beelink Mini PC for all image sizes. As explained in Section 3.1, these values indicate that the client side (MacBook Pro for experimental setup) was able to send the frames to the server side and receive them back with no observable lag. Time differences varied from a minimum of about 29 ms to a maximum of up to 35 ms, displaying a range of around 6ms. Table 2 shows the average and standard deviation of all time differences across all devices and resolutions. The average of all time differences between frames is 33.30 ms. As for the standard deviations, the average is 4.154 ms across all devices and resolutions. This standard deviation results from fluctuations of transfer time and frame processing time rather than frame capturing time variance in the camera.

**Table 2**. The time intervals between frames when a video stream is sent from client to server and then received back. Ideally the value of interval should match the frame rate of the video. The server was changed between desktop PC, Jetson Nano, and Mini PC.

| Video resolution | Desktop PC (ms) | Jetson Nano (ms) | Mini PC (ms) |
| --- | --- | --- | --- |
| 320x240 | 33.319 ± 3.892 | 33.326 ± 4.197 | 33.333 ± 3.930 |
| 640x480 | 33.255 ± 4.121 | 33.186 ± 5.255 | 33.320 ± 4.462 |
| 960x540 | 33.303 ± 3.722 | 33.299 ± 4.108 | 33.315 ± 3.932 |
| 1280x720 | 33.322 ± 3.984 | 33.303 ± 3.933 | 33.316 ± 3.867 |

In regards to the communication speed, there were no apparent issues with the throughput or delay across all platforms with a camera running at 30 fps. The communication speed is affected by the Wi-Fi performance of the devices in the client and server, wireless link quality, and the video frame processing time in client and server. Given that the wireless link will be inside of a vehicle and modern cellular phones that support high speed Wi-Fi such as 802.11 G/N/AC, we don't expect a shortage of speed for the current setting.

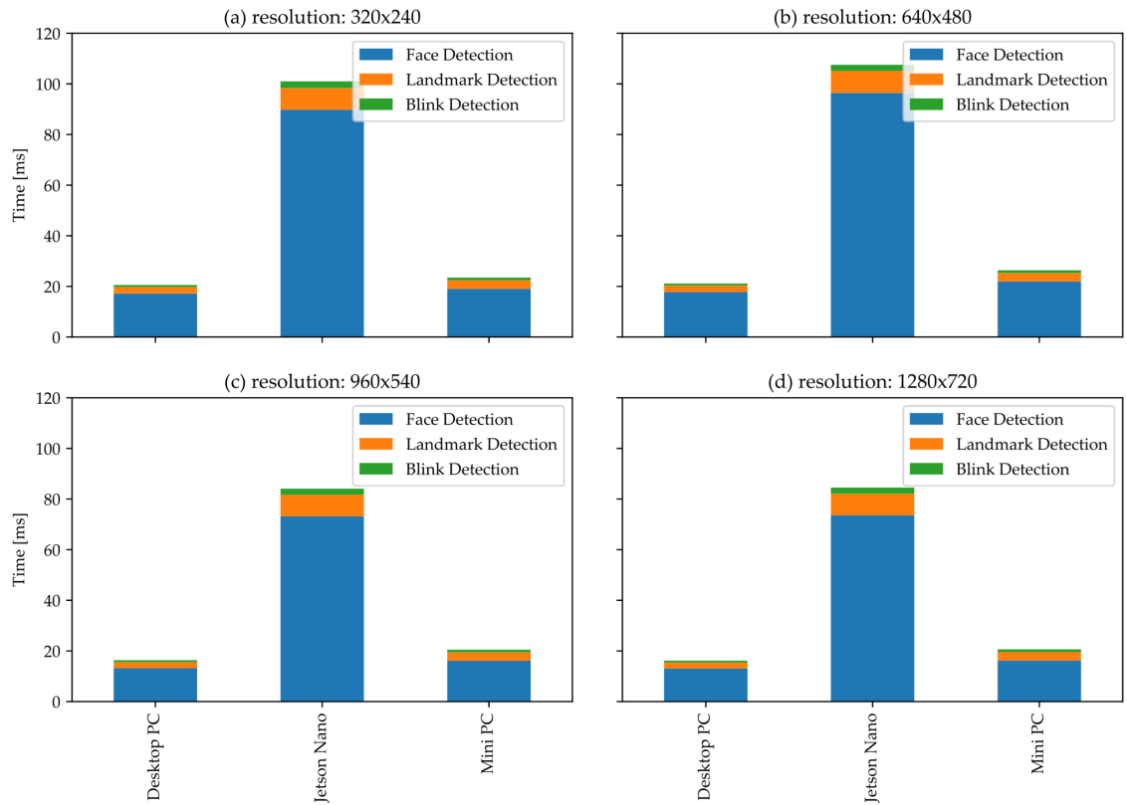

**Figure 12**. Inference time duration of face detection, landmark detection, blink detection, and total at various video resolutions. Across all video resolutions the Jetson Nano took four times more processing time than Desktop PC or Mini PC. The processing time in Desktop PC and Mini PC was smaller than 33ms meeting the 30 fps frame rate requirement.

Calculating the required communication throughput from the raw image size and frame rate, we could see that sending videoframes with resolution of 1280x720 at 30 fps requires a relatively higher range of WiFi speed. For example, multiplying 1280 x 720 (pixels) x 24 bit (RGB each 8 bit) x 30 fps leads to a link speed of 632 Mbps. 1280x720 was the maximum resolution and 30 fps was the maximum frame rate that we could use with the built-in Macbook Pro camera. However, when we send the videoframe through WebRTC we can choose to send frames in compressed format either VP8 or H264, which would still allow us to use the same WiFi throughput for higher frame rate and resolution for future uses. Certain blink features, such as eye opening velocity, can be analyzed more accurately with higher frame rates.

*4.2. Processing Time*

Figure 12 shows the various processing times of detecting the face, facial landmarks, and blinks at various video resolutions in the candidate servers. Across all video resolutions, the Jetson Nano typically took more than 4 times more time than the Desktop PC or Mini PC. The same data is shown in Table 3, where the average processing time for each operation on each device is displayed as well. In general, the average total processing time for the Windows 10 computer (Desktop PC) across all resolutions was 18.54 ms with an average standard deviation of 0.85 ms. The average total processing time for the Jetson Nano was 94.27 ms with an average standard deviation of 3.02 ms. Lastly, the average total processing time for the Beelink (Mini PC) was 22.73 ms with an average standard deviation of 1.56 ms.

As demonstrated by the data mentioned above, the Jetson Nano Computer took significantly longer to run the inference operations at each frame compared to other platforms. Even with the fastest operation time in Jetson Nano, in the case of 950x540 resolution, it took 84 ms to process one frame, which far exceeds the desirable 33 ms value for the 30 fps camera video frame rate. Thus, we choose not to use Jetson Nano for our drowsiness detection system because an operation time longer than its frame rate will accumulate as lag over the time, making it impossible to alert drivers in real time. The Windows 10 computer performed inference operations in the shortest amount of time, but the size of the computer and the required power to run the computer makes it impractical for use in vehicles. In the case of the Mini PC, the operation time meets the frame rate requirement. Even the longest operation time took 26.364 ms, which was less than 33 ms. The mobile size of the Mini PC and its efficient power consumption (28 W at maximum) makes it possible to apply this system in vehicles. Therefore, we concluded that the Mini PC is best suited to run our inference operations, which include Face Detection, Landmark Detection, and Blink Detection. Notably, predicting drowsiness values using the analyzed blinks is also a significant step of the drowsiness detection algorithm, but since this step is not performed at every frame and only when a new blink is detected, we did not measure the processing time.

In reality, there are 4 processes in drowsiness detection: face, landmark, blink, drowsy. The first three operations must happen at every frame, while the last one only needs to happen every blink, which happens in much larger time intervals, which is less than 20 time the minute. Thus, we didn't conduct the processing time analysis for the drowsiness detection algorithm itself.

**Table 3**. Inference time duration of face detection, landmark detection, blink detection, and total at various video resolutions. Desktop PC and Mini PC show similar performance while at Jetson Nano the processing time is longer. Units are in milliseconds.

| Inference | Video Resolution | Desktop PC (ms) | Jetson Nano (ms) | Mini PC (ms) |
|---|---|---|---|---|
| Face Detection | 320x240 | 17.177 ± 0.816 | 89.799 ± 1.064 | 19.031 ± 1.368 |
|  | 640x480 | 17.752 ± 0.925 | 96.453 ± 2.048 | 21.935 ± 1.928 |
|  | 960x540 | 13.165 ± 0.762 | 73.182 ± 2.282 | 16.078 ± 1.169 |
|  | 1280x720 | 13.082 ± 0.571 | 73.585 ± 2.288 | 16.218 ± 1.152 |
| Landmark Detection | 320x240 | 2.543 ± 0.121 | 8.533 ± 0.348 | 3.410 ± 0.332 |
|  | 640x480 | 2.570 ± 0.283 | 8.633 ± 0.590 | 3.346 ± 0.651 |
|  | 960x540 | 2.391 ± 0.116 | 8.503 ± 0.314 | 3.375 ± 0.369 |
|  | 1280x720 | 2.319 ± 0.320 | 8.528 ± 0.221 | 3.362 ± 0.383 |
| Blink Detection | 320x240 | 0.835 ± 0.063 | 2.645 ± 5.164 | 1.037 ± 0.124 |
|  | 640x480 | 0.803 ± 0.065 | 2.408 ± 0.106 | 1.083 ± 0.170 |
|  | 960x540 | 0.790 ± 0.316 | 2.398 ± 0.108 | 1.019 ± 0.128 |
|  | 1280x720 | 0.745 ± 0.037 | 2.401 ± 0.069 | 1.035 ± 0.125 |
| Total | 320x240 | 20.554 ± 0.876 | 100.997 ± 5.317 | 23.478 ± 1.533 |
|  | 640x480 | 21.126 ± 1.042 | 107.524 ± 2.121 | 26.364 ± 2.018 |
|  | 960x540 | 16.345 ± 0.817 | 84.083 ± 2.344 | 20.471 ± 1.306 |
|  | 1280x720 | 16.146 ± 0.671 | 84.514 ± 2.305 | 20.615 ± 1.366 |

We found that the face detection step is the most demanding operation compared to the landmark detection and blink detection steps. The face detection operation took more than 80% of the total processing time for all resolutions across all devices. The face detection processing time also varied depending on the resolution, not necessarily proportional to the image size. For example in Table 3, 960x540 and 1280x720 resolutions took smaller time than 320x240 or 640x480 resolutions. This may be due to differences in the image scaling process that the program performs before inputting the image to the face detection model. Depending on the specific image size, the scaling operation could simply involve increasing the pixel intensity, while for their sizes may involve complex fraction multiplications. For landmark and blink detection, we did not observe changes due to image resolution.

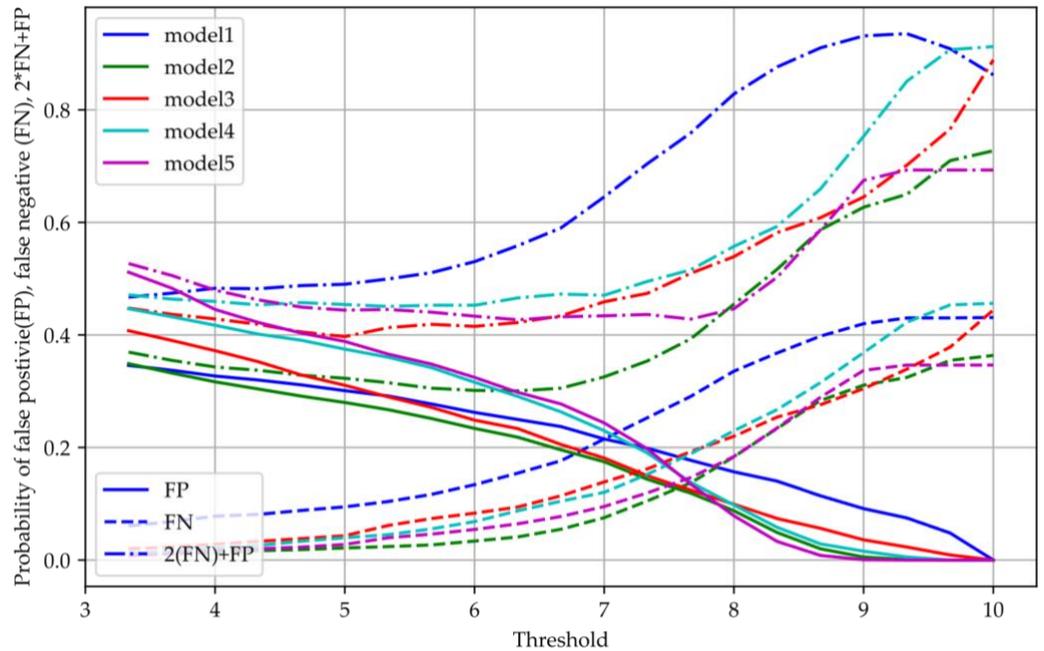

**Figure 13**, The false positive (FP), false negative (FN), and cost function of 2FN+FP values are shown across a range of threshold values, 3.3 to 10. Each color represents different models (1-5) generated from the cross-validation process as elaborated in section 2.3.3. Optimal threshold can be determined by finding a threshold that minimizes 2FN+FP. This approach solves the trade-off problem between false positive and false negative rates.

At a frame rate of 30 fps, the processing can be done in time on both a Mini PC and desktop PC. However, if we increase the frame rate to 60 fps in future, the processing needs to happen within 16.6 ms, and only a Desktop PC with image resolutions of 960x540 or 1280x720 would barely meet this requirement. In such cases, we think extra hardware such as the Google Coral stick in combination with the Mini PC could be used to improve the inference speed of facial detection.

*4.3 Threshold Optimization*

To display the relationship between threshold values and false negative (FN) and false positive (FP) rates, we calculated the false negative and false positive rates of each model for each threshold value from 3.33 to 10.0 in increments of 0.33 and graphed as shown in Figure 13.

As shown in Figure 13, the FP rate tends to decrease as the threshold increases because a higher threshold leads to a greater confidence of drowsiness detection. The solid curves, which represent FP curves, display an FP rate between 0.3 and 0.5 when the threshold is 3.33 but decrease to 0.1 when the threshold is approximately 0.8. On the contrary, the FN rate tends to increase as the threshold increases because there are more instances falsely detected as negative when the threshold gets higher. The dashed curves, which represent the FN rates, display an FN rate close to 0 when the threshold is 3.3 but increases to around 0.4 when the threshold is 10. The two different tendencies create a trade-off situation when choosing a threshold value. Thus, we created another metric that weighs the sensitivity and specificity differently. For example, 2FN + FP will penalize FN twice more than FP. The dotted-dashed curve, which represents the 2FN+FP, displays slight dips at relatively low thresholds before increasing at a fast rate. Except for model 1, the 2FN+FP value tends to stay between 0.3 and 0.5 until the threshold is 7, then rises rapidly from 0.5 to 0.9 as the threshold increases. Since we wanted to find a threshold that minimizes 2FN + FP, where minimizing the FN value is more important, the threshold will be between 3.3 and 7. Flat lines throughout the curves indicate that differences in the threshold do not affect the accuracy of the models at those regions.

**Table 4**. Two sets of 2FN+FP, FP, and FN values are listed: the first set shows these values for the optimal threshold values and the second show these values for the default threshold value (6.67). When the optimal threshold differs from the default threshold (as in models 3 and 4), the FN value is minimized at the expense of an increased FP value.

| Model number | 2FN+FP at optimal threshold | FP at optimal threshold | FN at optimal threshold | Optimal threshold | 2FN+FP at default threshold | FP at default threshold | FN at default threshold |
|---|---|---|---|---|---|---|---|
| 1 | 0.47 | 0.35 | 0.061 | 3.33 | 0.59 | 0.24 | 0.177 |
| 2 | 0.30 | 0.22 | 0.041 | 6.33 | 0.31 | 0.20 | 0.054 |
| 3 | 0.40 | 0.31 | 0.043 | 5.00 | 0.43 | 0.21 | 0.114 |
| 4 | 0.45 | 0.36 | 0.055 | 5.33 | 0.47 | 0.26 | 0.104 |
| 5 | 0.43 | 0.30 | 0.064 | 6.33 | 0.43 | 0.28 | 0.077 |

Model 1 displays consistently poorer performance than the other models. This can be seen from the higher values of blue solid line, blue dashed line, and blue dotted-dashed line. We predict that incorrect labels of some of the videos specifically in the test set of Model 5 could have caused this decrease in performance. Notably, this significant dip in performance will be taken into account when calculating the final drowsiness value, which is the weighted average of all of the drowsiness classifications. We further describe this step in Section 2.3.3.

Table 4 summarizes the minimum values of 2FN+FP, FP, and FN for each model. In addition, we compared the values at the optimized threshold to those at the default threshold (6.67). With the optimal thresholds, the minimum values of FN were smaller, and values of FP were larger compared to those with default threshold. In the table, Model 2 and Model 5's optimal thresholds are both 6.33, which is very close to default 6.6, and the values of FP and FN are similar for both thresholds. However, for Model 3 and Model 4 which have different optimal thresholds than the default threshold, the FN rates are 63.3%, 48.1% and lower at the expense of increased FP rates of 47.6% and 38.5% for model 3 and model 4 respectively. Thus, the threshold optimization algorithm will reduce the number of false negative cases at the expense of increasing the number of false positive cases. Thus, the drivers can almost always be alerted of their drowsiness, despite occasionally being annoyed by false alarms.

We chose a weight of 2 for FN arbitrarily to penalize FP more than FN. However, a more precise weight can be assigned through additional research on drivers' experiences and how they feel about the false alarms versus missed alerts when drowsy. This effort as a future work will require both quantitative and qualitative analysis for improving both the driving experience and safety.

*4.4 Future work*

The dataset we used for our drowsiness detection algorithm was the one used by Ghoddoosian et al., which is composed of 10-minute recordings made by 60 people for each of the alert, low-vigilant, and drowsy states. We believe that since drowsiness behaviors differ from person to person, the decoding accuracy of the model can be greatly improved by personalizing the model for each user (training the model with a dataset consisting solely of the user's videos).

Furthermore, for our Threshold Optimization and Voting algorithm, we assigned a weight of two for the TF and FN values respectively. We arbitrarily chose the value of two simply for the purpose of our experiment and think further research can be conducted to find the optimal ratio of the weight of TP to TN or FN to FP values in finding the optimal threshold.

**5. Conclusion**

In this paper, we proposed and developed an embedded system that allows a neural network-based drowsiness detection model to run in real-time in vehicles and discovered that the most practical system setups of a Beelink Mini PC as the server and a phone as the web app client. The drowsiness detection algorithm can run smoothly with no accumulating lag time. For the Beelink Mini PC, the total processing time for all drowsiness detection processes is 22.73 ms, which is less than the maximum 33 ms value required to process frames of a 30 fps camera video stream. We have also shown that communication throughput between the client and server was adequate to send video images between the two. Both the measured average and the median time interval between frames were around 33.33 ms. Thus, we have implemented a system that can run drowsiness prediction effectively via a client and server platform. We also developed an algorithm that calculates an optimal threshold value considering the trade-off between the safety and the convenience of the user. In some models, the algorithm reduced the false negative rates by 63.3% at the expense of increasing the false positive rate by 47.6%.

In the future, a personalized drowsiness model (trained based on a dataset of a single person) can be developed to improve the accuracy in comparison to the existing model used in this paper, which was trained with a dataset of 60 participants. Our Threshold Optimization Algorithm can be improved further by exploring different cost functions which optimize the ratio between the FN rate and FP rate. Overall, we believe that our drowsiness prediction platform can be used by other researchers aiming to create a real-time embedded system implementation of drowsiness prediction in vehicles.

**Acknowledgments:** We would like to thank Mr. Jared Lera and the Applied Computing Foundation for providing guidance throughout our project and for suggesting Ghoddoosian's model as a foundation for our drowsiness detection system. Also, we would like to thank Professor Vassilis Athitsos from the University of Texas at Arlington for guiding us on the structure and submission process of this paper.